\documentclass[runningheads]{llncs}
\usepackage[T1]{fontenc}
\usepackage{algorithm2e}
\RestyleAlgo{ruled}
\usepackage{graphicx}
\usepackage{amsmath}
\usepackage{subfigure}
\begin{document}
\title{Optimizing Federated Learning for Medical Image Classification on Distributed Non-iid Datasets with Partial Labels}
%
\titlerunning{FedFBN}
%
\author{Pranav Kulkarni\inst{1} \and
Adway Kanhere\inst{1,2} \and
Paul H. Yi\inst{1} \and
Vishwa S. Parekh\inst{1}}
%
\authorrunning{P. Kulkarni et al.}
%
\institute{University of Maryland Medical Intelligent Imaging (UM2ii) Center, Baltimore, MD \\
\email{\{pkulkarni,akanhere,pyi,vparekh\}@som.umaryland.edu} \and
Department of Biomedical Engineering, Johns Hopkins University, Baltimore, MD \\
\email{akanher1@jhu.edu}}
%
\maketitle              
\begin{abstract}
Numerous large-scale chest x-ray datasets have spearheaded expert-level detection of abnormalities using deep learning. However, these datasets focus on detecting a subset of disease labels that could be present, thus making them distributed and non-iid with partial labels. Recent literature has indicated the impact of batch normalization layers on the convergence of federated learning due to domain shift associated with non-iid data with partial labels. To that end, we propose FedFBN, a federated learning framework that draws inspiration from transfer learning by using pretrained networks as the model backend and freezing the batch normalization layers throughout the training process. We evaluate FedFBN with current FL strategies using synthetic iid toy datasets and large-scale non-iid datasets across scenarios with partial and complete labels. Our results demonstrate that FedFBN outperforms current aggregation strategies for training global models using distributed and non-iid data with partial labels.

\end{abstract}

\section{Introduction}
Federated Learning (FL) is a promising machine learning paradigm that allows the training of machine learning algorithms, such as deep neural networks, on multiple distributed datasets while preserving privacy \cite{fl_digitalhealth,fl_medicine} The FL approach involves a central server that trains a global model by iteratively aggregating knowledge from local models that are trained on different datasets stored on separate nodes. Specifically, each node trains its local model on its dataset and communicates the model to the server, which intelligently aggregates the local models into a single global model for the next round of training. In the field of medical imaging, FL has enabled the training of large-scale deep learning models using homogeneous data from multiple institutions without sharing sensitive patient data. However, most medical imaging datasets are heterogeneous, acquired from different domains with partial and incomplete annotations. This poses a significant challenge for the FL approach, as the local models may differ significantly from each other, leading to suboptimal aggregation of knowledge and decreased accuracy of the global model. 

Batch normalization (BN) layers are a core element in most deep learning model architectures. By normalizing, recentering, and rescaling the layer inputs, BN ensures rapid and stable model convergence by learning a moving mean of the training data statistics \cite{ioffe2015batch}. However, recent literature has indicated that, due to domain shift, BN layers significantly impact the convergence of FedAvg, the de-facto strategy for weight aggregation in a FL setup \cite{li2021fedbn}. As a result of domain shift, there is a disconnect between the statistics learned by BN layers at each local model, and thus na\"ively averaging these statistics has drastically reduced model accuracy \cite{li2021fedbn}. Furthermore, FedAvg has shown to be affected by catastrophic forgetting during domain adaption tasks \cite{liang2020think}.

To solve the problem of suboptimal aggregation from distributed datasets, FedBN has recently emerged as a popular technique \cite{li2021fedbn}. Compared to other FL algorithms, such as FedAvg \cite{fedavg} and FedProx \cite{li2020federated}, FedBN has demonstrated success in creating a global model that can generalize well to non-iid data with domain shift. This is achieved by avoiding the aggregation of BN layers during communication rounds. FedBN has been shown to improve the performance of FL models on non-iid data and reduce the negative impact of data heterogeneity. However, FedBN is limited to training personalized local models rather than global models. Recent literature has demonstrated that, in the case of partial labels when participating datasets have either different of partially overlapping labels, FedBN converges to a suboptimal solution and exhibits poor performance \cite{wang2023batch}. Meanwhile, in a non-federated setup, transfer learning (TL) has emerged as a popular technique to rapidly adapt pretrained networks to medical image classification tasks. While the overall effectiveness of TL on model performance is unclear \cite{raghu2019transfusion} due to a significant domain shift from the source to target data in addition to feature duplication, other approaches have demonstrated the importance of fine-tuning BN layers to achieve greater model accuracy using TL \cite{li2016revisiting}. To that end, we propose FedFBN, a federated learning framework that draws inspiration from FedBN and transfer learning by using pretrained networks as the model backend and freezing the batch normalization layers throughout the collaborative learning phase. We evaluated the FedFBN framework for aggregating knowledge from distributed iid and non-iid datasets with complete and partial labels and compared its performance to individual node baseline models, models trained on centrally aggregated data, and different FL aggregation strategies, including FedAvg and FedBN. 

\section{Methods}

\subsection{FedFBN}

The FedFBN algorithm is implementated as a modification of the FedBN algorithm \cite{li2021fedbn}. Rather than avoiding the aggregation of the BN layer statistics, FedFBN keeps these statistics frozen i.e. avoids updating them in the training process. The FedFBN algorithm is described in detail in Algorithm \ref{alg:fedfbn}.

\begin{algorithm}
\SetAlgoLined
\caption{FedFBN}
\label{alg:fedfbn}
\KwIn{$K$ clients indexed by $k$, model layer indexed by $l$, initialized model parameters $w_{0,k}^{(i)}$, training epochs before aggregation $E$, and total federated rounds $N$;}
\For{$\text{each round}\ n=1,2,...,N$} {
  \For{$\text{each client}\ k\ \text{and each layer}\ l$} {
    \If{$\text{layer}\ l\ \text{is not BatchNorm}$} {
      $w_{n,k}^{(l)} \gets SGD(w_{n-1,k}^{(l)})$
    }
  }
  \If{$\text{mod}(t, E) = 0$} {
    \For{$\text{each client}\ k\ \text{and each layer}\ l$} {
      $w_{n,k}^{(l)} \gets \frac{1}{K}\sum_{i=1}^{K} w_{n,i}^{(l)}$
    }
  }
}
\end{algorithm}

\subsection{Datasets}

\textbf{NIH Chest X-Ray 14:}
The NIH dataset, curated by \cite{wang2017chestx}, consists of 14 disease labels with $N=112,120$ frontal-view CXRs from $M=30,805$ patients. We randomly divided the dataset into training (70\%, $N=78,571$, $M=21,563$), validation (10\%, $N=11,219$, $M=3,080$), and testing (20\%, $N=22,330$, $M=6,162$) splits while ensuring no patient appears in more than one split. Furthermore, we sampled two synthetic iid toy datasets, NIH (1) and (2), from the NIH dataset by equally distributing patients diagnosed with at least one positive finding with healthy patients to prevent patient overlap. Both synthetic datasets consisted of the same number of patients with distinct training ($M=10,781$) and validation ($M=1,540$) sets sampled from the global splits, while ensuring there is no patient overlap.

\textbf{CheXpert:}
The Stanford CheXpert dataset \cite{irvin2019chexpert} consists of 13 disease labels (7 shared with NIH) and an additional label for `normal' with $N$=224,316 CXRs from $M$=65,240 patients. To handle uncertain labels in the dataset, we chose the U-Zeros approach i.e. treating all uncertain labels as negatives. We randomly divided the dataset into training (70\%, $N=155,470$, $M=45,178$), validation (10\%, $N=22,736$, $M=6,454$), and testing (20\%, $N=45,208$, $M=12,908$) splits while ensuring no patient appears in more than one split. 

\textbf{MIMIC-CXR-JPG}:
To evaluate the generalizability of our methods, we utilized the MIMIC-CXR-JPG dataset \cite{goldberger2000physiobank,johnson2019mimic} as our external test set. Similar to the CheXpert dataset, the MIMIC dataset consists of 13 disease labels (7 shared with NIH) and an additional label for `normal' with $N=377,110$ CXRs from 65,379 patients. To handle uncertain labels in the dataset, we chose the U-Zeros approach i.e. treating all uncertain labels as negatives.

\subsection{Federated Learning Setup}

For our federated learning setup, we utilized a two-node surgical aggregation \cite{kulkarni2023surgical} setup to train a multi-label global meta-model. In short, surgical aggregation is a task-agnostic, semi-supervised framework for aggregating knowledge from distributed datasets with partial labels, consisting of a representation and task block. Due to it's agnostic nature, surgical aggregation serves as a backbone in our analysis. By changing the aggregation strategy used for aggregation weights for the representation block, we can evaluate various FL strategies including FedFBN. For our model architecture, we used DenseNet121 \cite{densenet}, initialized with ImageNet \cite{deng2009imagenet} weights. 

Prior to federation, the classification heads of all local models were warmed-up using transfer learning with learning rate of 1e-3. For our analysis, we modified the strategy used for weight aggregation in the representation block. For preprocessing, all images were downsampled to 224x224, normalized between 0 and 1, and augmented. We trained all models for 100 epochs with 1 epoch between two consecutive communication rounds ($E=1$) with a batch size of 64 and learning rate of 1e-5 using TensorFlow 2.8.1. At the end of each local training round, the binary cross entropy (BCE) losses at each node were communicated back to the server and the best performing global model was determined by the model with the smallest mean loss. 

\subsection{Experimental Setup}

We conducted four experiments that evaluate FedFBN across non-iid and iid datasets with partial and complete label distribution across two nodes. Across all four experiments, we will compare the performance of models training using FedFBN with FedAvg and FedBN. Due to the inability of FedBN to learn a global model, we measured the performance of each local FedBN model. Additionally, we will also compare with baseline and centrally aggregated models. Here, baseline refers to models trained with the complete dataset, while centrally aggregated models were locally trained by na\"ively concatenating each dataset without harmonizing disease labels. Furthermore, prior literature has demonstrated that utilizing partial loss to tackle partial labels in a federated setup yields poor performance on unseen labels \cite{kulkarni2023surgical}.

For all experiments, as model metrics, we measured BCE loss and the average area under receiver operating characteristic curve (AUROC) score. In a multi-label scenario, the AUROC score is defined as the mean AUROC score across all disease labels. We compared the performance of FedFBN with other FL strategies and baseline models using bootstrapping and two-tailed paired t-test. We calculated the 95\% confidence interval (CI) for the AUROC scores and compared them using bootstrapping and a paired two-tailed t-test. Statistical significance was defined as $p < 0.05$.

\begin{enumerate}
    \item \textbf{Distributed iid data with complete labels:} 
    This experiment encapsulates the best possible case for a federated learning setup. While the data is distributed, it is iid and contains the complete representation of all labels. We use the two synthetic NIH (1) and (2) datasets to evaluate the performance of FedFBN. Since both datasets contain the complete label distribution, the baseline and centrally aggregated models are equivalent. We evaluated all models on the held-out NIH test set and the MIMIC external test set.
    
    \item \textbf{Distributed iid data with partial labels:} 
    In this experiment, we add a layer of complexity to the problem due to the presence of partial labels. To simulate this case, we randomly pruned the disease labels from the two synthetic NIH (1) and (2) datasets to yield datasets containing 11 and 7 disease labels, with 4 labels shared across both. We evaluated all models on the held-out NIH test set by measuring the model performance on the 4 shared labels as well as each subset of the label distribution. Similarly, we also evaluated all models on the MIMIC external test set.
    
    \item \textbf{Distributed non-iid data with complete labels:} 
    The complexity of the problem increases drastically as we switch from iid to non-iid data. Due to domain shift, the features learned by each local model differ. In this experiment, we train a global model using the entire NIH and CheXpert datasets on the 7 shared labels using surgical aggregation with differential learning rates of 1e-5 and 5e-5 to tackle domain shift. As a result, we can effectively evaluate across all three large-scale datasets in our analysis. We evaluate all models on the held-out NIH and CheXpert datasets, along with the MIMIC external test set on the 7 shared labels. Additionally, we cross-evaluate the NIH baseline model on the CheXpert test set and vice versa.
    
    \item \textbf{Distributed non-iid data with partial labels:} 
    Out of the all scenarios considered in our experimental setup, tackling distributed non-iid data with partial labels is the most complex problem due to domain shift and the presence of unseen disease labels. Here, we train a global model using the entire NIH and CheXpert datasets using surgical aggregation with differential learning rates of 1e-5 and 5e-5 to tackle domain shift. This yields us a 20 disease label classifier and we can effectively evaluate across all three large-scale datasets. We evaluate all models on the held-out NIH and CheXpert datasets, along with the MIMIC external test set on the 7 shared labels and all observed disease labels in each dataset. Additionally, we cross-evaluate the NIH baseline model on the CheXpert test set and vice versa.
\end{enumerate}

\section{Results}

In the experiment with distributed iid data with complete labels, on the held-out NIH test set, we observe that the performance of FedFBN is comparable to, if not slightly better than, the baseline NIH model with a mean AUROC 0.82 ($p=0.07$). We also observe that FedFBN performs comparably to FedAvg ($p=0.25$), while slightly outperforming the local FedBN model ($p=0.01$). On the MIMIC external test set, we observe that FedFBN generalizes better to unseen distributions and outperforms all models with a mean AUROC of 0.75 ($p<0.001$). Results are detailed in Table \ref{tab:exp4_results}.

\begin{table}[!htb]
    \centering
    \caption{Model metrics for iid datasets with complete labels on held-out NIH test set and external MIMIC test set. 95\% CI for mean AUROC scores provided in brackets. Best performing models are highlighted in bold.}
    \includegraphics[width=0.7\linewidth]{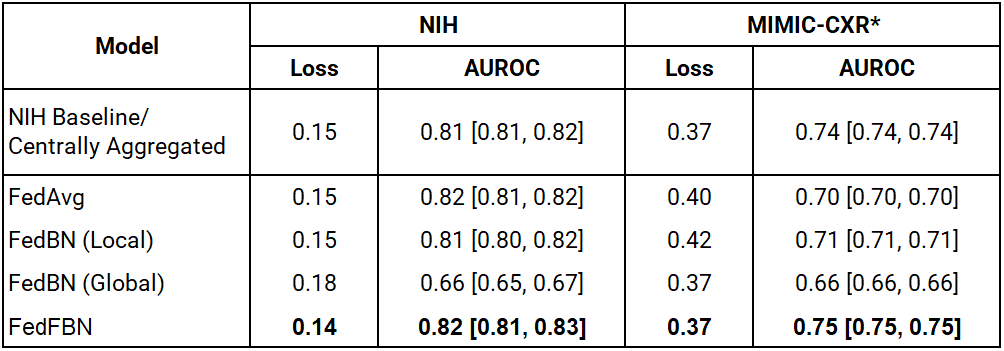}
    \label{tab:exp4_results}
\end{table}

When switching to a partial label distribution with iid data, on the held-out NIH test set, we observe that FedFBN performs comparable to the baseline NIH model across the each subset of the label distribution with a mean AUROC of 0.82 ($p=0.22$) for NIH (1) and 0.85 ($p=0.11$) for NIH (2), and all 14 NIH disease labels with a mean AUROC of 0.82 ($p=0.42$). Furthermore, on the 4 shared labels between both synthetic datasets, FedFBN outperforms the baseline model with a mean AUROC of 0.89 ($p<0.001$). Moreover, across all test sets, FedFBN outperforms FedAvg and the local FedBN model ($p<0.001$). As expected, the local FedBN model performs poorly due it's drawback in scenarios with partial labels. We observe similar results on the MIMIC external test set, where FedFBN generalizes better and outperforms all models with a mean AUROC of 0.75 ($p<0.001$). Results are detailed in Table \ref{tab:exp3_results}.

\begin{table}[!htb]
    \centering
    \caption{Model metrics for iid datasets with partial labels on held-out NIH test set and external MIMIC test set. 95\% CI for mean AUROC scores provided in brackets. Best performing models are highlighted in bold.}
    \subfigure[4 Shared Disease Labels]{\includegraphics[width=0.4\linewidth]{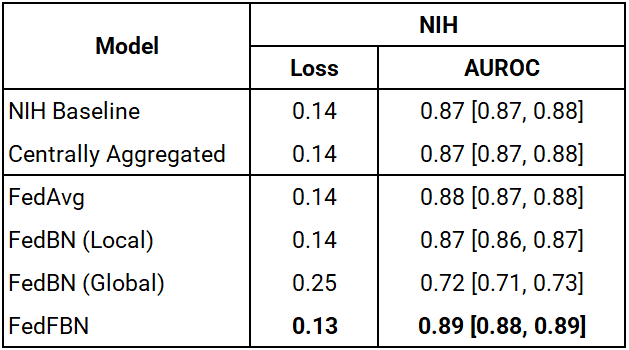}} \\
    \subfigure[All Disease Labels]{\includegraphics[width=\linewidth]{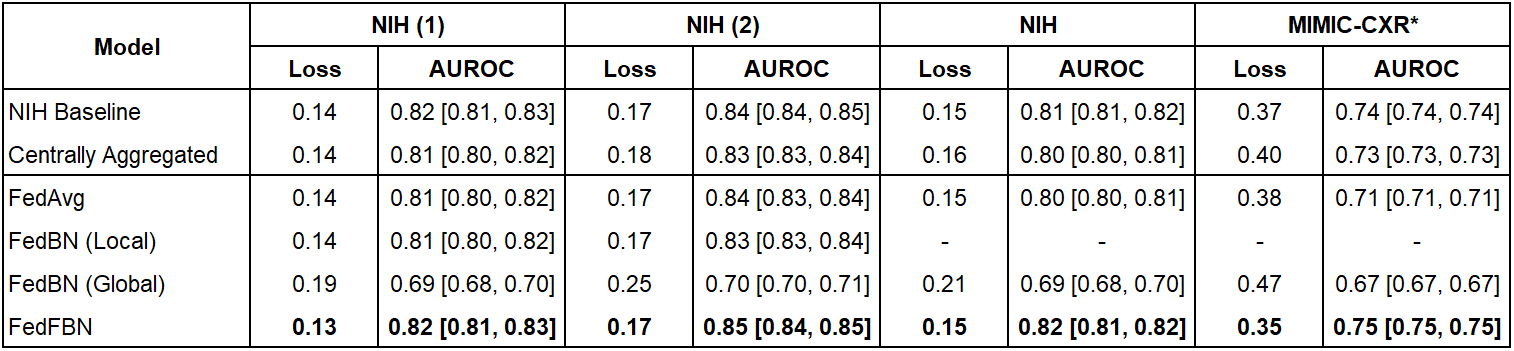}}
    \label{tab:exp3_results}
\end{table}

Despite the added complexity of distributed non-iid data with complete disease labels, we observe that FedFBN performs comparably to the baseline NIH model with a mean AUROC of 0.84 ($p=0.1$) and the baseline CheXpert model with a mean AUROC of 0.77 ($p=0.98$). We finally observe the drawbacks of FedAvg in this experiment as it performs worse than the centrally aggregated model and is outperformed by FedBN. However, on the NIH test set, FedFBN outperforms the local FedBN model ($p=0.01$). On all other evaluations across both test sets, FedFBN outperforms FedAvg and the local FedBN model ($p<0.001$). We observe similar results on the MIMIC external test set, where FedFBN generalizes better and outperforms all models with a mean AUROC of 0.81 ($p<0.001$). Results are detailed in Table \ref{tab:exp2_results}.

\begin{table}[!htb]
    \centering
    \caption{Model metrics for non-iid datasets with complete labels on held-out NIH and CheXpert test sets and external MIMIC test set. 95\% CI for mean AUROC scores provided in brackets. Best performing models are highlighted in bold.}
    \includegraphics[width=\linewidth]{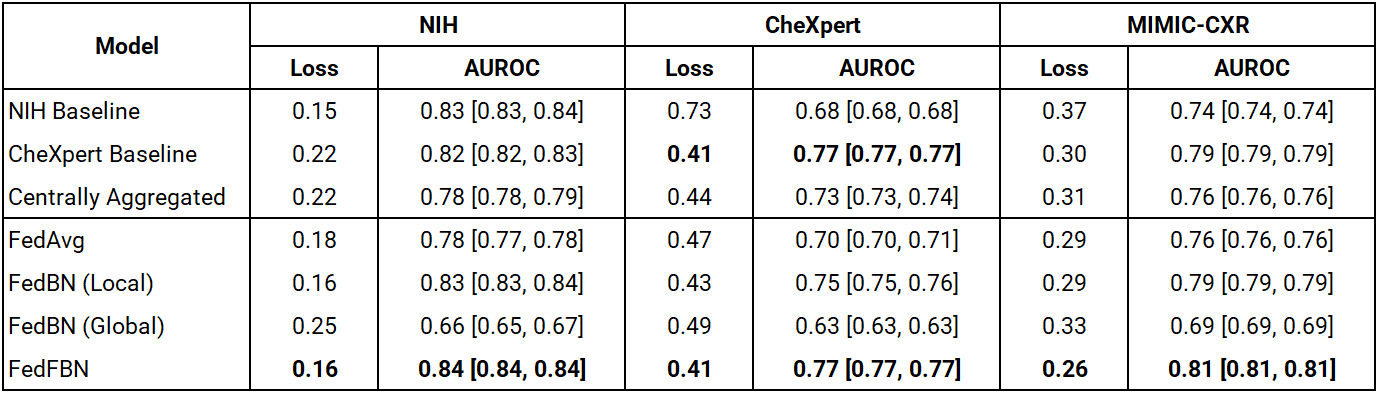}
    \label{tab:exp2_results}
\end{table}

In the final experiment, with the highest complexity due to distributed non-iid data with partial labels and domain shift, we yet again observe FedFBN's superiority over other aggregation strategies. When the 7 shared disease labels are considered, FedFBN performs better than the baseline NIH model with a mean AUROC of 0.84 ($p=0.03$) while performing comparably to the baseline CheXpert model with a mean AUROC of 0.77 ($p=0.98$). Furthermore, FedFBN outperforms FedAvg and the local FedBN model across both test sets ($p<0.001$). When all observed disease labels across both test sets are considered, we observe that FedFBN performs comparable to the baseline NIH model with a mean AUROC of 0.81 ($p=0.06$), but outperforms the baseline CheXpert model with a mean AUROC of 0.75 ($p<0.001$). We observe that apart from performing comparably to FedBN on the NIH test set ($p=0.72$), FedFBN outperforms both FedAvg and the local FedBN model ($p<0.001$). We observe similar results on the MIMIC external test set, where FedFBN generalized better and outperforms all models with a mean AUROC of 0.75 ($p<0.001$). Results are detailed in Table \ref{tab:exp1_results}.

\begin{table}[!htb]
    \centering
    \caption{Model metrics for non-iid datasets with partial labels on held-out NIH and CheXpert test sets and external MIMIC test set. 95\% CI for mean AUROC scores provided in brackets. Best performing models are highlighted in bold.}
    \subfigure[7 Shared Disease Labels]{\includegraphics[width=\linewidth]{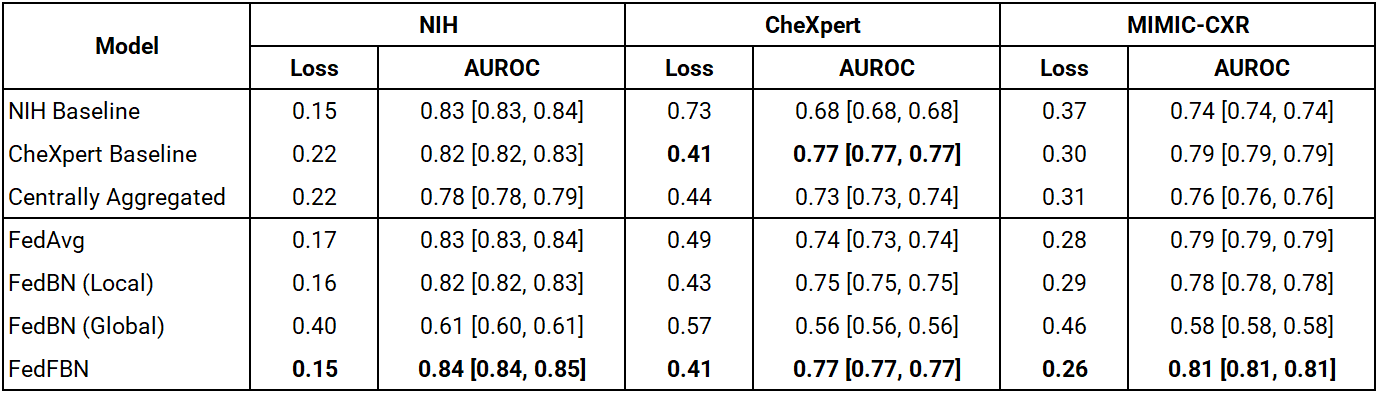}} \\
    \subfigure[All Disease Labels]{\includegraphics[width=\linewidth]{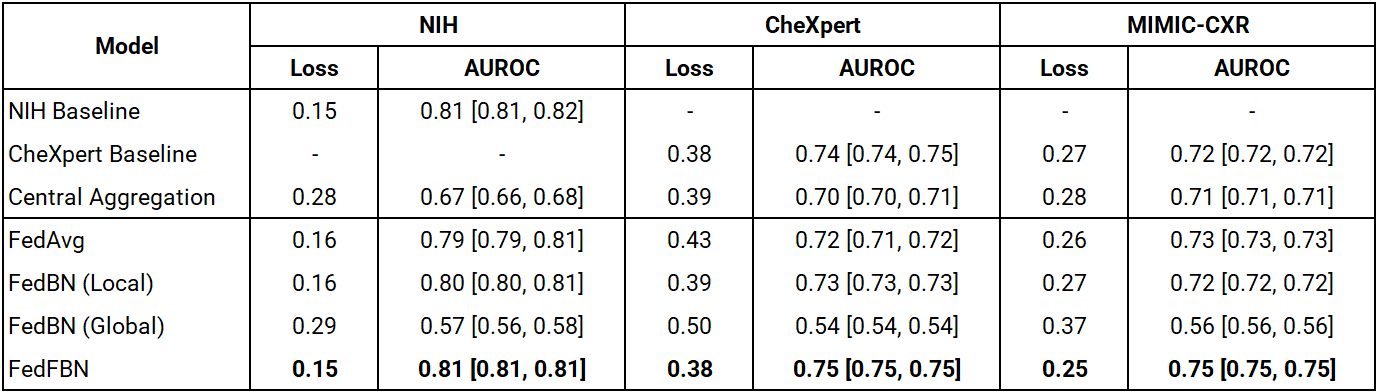}}
    \label{tab:exp1_results}
\end{table}



\section{Discussion}

The distributed nature of medical imaging data has impacted the clinical utility of deep learning models. As a result, tackling distributed non-iid datasets with partial labels is a crucial step towards training clinically useful models. Our results demonstrate that FedFBN outperforms current FL aggregation strategies for training global models, while generalizing better to unseen distributions and performing comparably to baseline models. Despite FedAvg being the de-facto standard for FL setups, our work reinforces its limitations with non-iid data. Furthermore, our work also highlights the limitations in FedBN, another popular choice among FL setups dealing with non-iid data. While FedBN addresses many drawbacks with FedAvg, its limitation to training personalized local models rather than global models drastically affects its utility in the clinical workflow. While our work is limited to smaller-scale experiments that do not reflect a realistic FL setup with numerous nodes, our results indicate the superiority of FedFBN over FedAvg and FedBN and demonstrate that FedFBN has the potential to train clinically useful models by leveraging large-scale centrally trained networks to address the negative impact of batch normalization due to domain shift with distributed non-iid datasets with partial labels. 

\bibliography{references}

\end{document}